\documentclass[sigconf]{acmart}
\usepackage{graphicx}
\usepackage{textcomp}
\usepackage{url}
\usepackage{xcolor,colortbl}
\usepackage{float}
\usepackage{subfigure}
\usepackage{makecell}
\usepackage{multirow}
\usepackage{soul}
\usepackage{algorithm}
\usepackage{algpseudocode}
\usepackage[algo2e]{algorithm2e}
\usepackage{pifont}
\usepackage{booktabs}
\usepackage{ulem}
\usepackage{pifont}
\usepackage{comment}
\usepackage{enumerate}
\AtBeginDocument{%
  \providecommand\BibTeX{{%
    \normalfont B\kern-0.5em{\scshape i\kern-0.25em b}\kern-0.8em\TeX}}}

\definecolor{Color0}{RGB}{0,0,252}
\definecolor{Color1}{HTML}{fff44f} 
\definecolor{Color2}{HTML}{d3d3d3} 
\definecolor{Color3}{HTML}{b9ffda}

\copyrightyear{2024} 
\acmYear{2024} 
\setcopyright{acmcopyright}\acmConference[ICCAD '24]{IEEE/ACM International
Conference on Computer-Aided Design}{}{New Jersey,
NJ, USA}
\acmBooktitle{IEEE/ACM International Conference on Computer-Aided Design (ICCAD
'24), October 27-31, 2024, New Jersey,
NJ, USA}
\acmPrice{15.00}
\acmDOI{10.1145/3508352.3549379}
\acmISBN{978-1-4503-9217-4/22/10}

\begin{document}

\title{AyE-Edge: Automated Deployment Space Search Empowering \uline{A}ccuracy \uline{y}et \uline{E}fficient Real-Time Object Detection on the \uline{Edge}}

\author{Chao Wu}
\authornote{Both authors contributed equally to this research.}
\author{Yifan Gong}
\authornotemark[1]
\affiliation{%
  \institution{Northeastern University}
  \city{Boston}
  \state{MA}
  \country{USA}
}
\author{Liangkai Liu}
\authornotemark[1]
\affiliation{%
  \institution{University of Michigan}
  \city{Ann Arbor}
  \state{MI}
  \country{USA}
}

\author{Mengquan Li}
\authornote{Corresponding author.}
\affiliation{%
  \institution{Hunan University}
  \city{Hunan}
  \country{China}}
\email{mengquanli@hnu.edu.cn}

\author{Yushu Wu}
\affiliation{%
  \institution{Northeastern University}
  \city{Boston}
  \state{MA}
  \country{USA}
}

\author{Xuan Shen}
\affiliation{%
  \institution{Northeastern University}
  \city{Boston}
  \state{MA}
  \country{USA}
}

\author{Zhimin Li}
\affiliation{%
  \institution{Northeastern University}
  \city{Boston}
  \state{MA}
  \country{USA}
}

\author{Geng Yuan}
\affiliation{%
  \institution{University of Georgia}
  \city{Athens}
  \state{GA}
  \country{USA}
}

\author{Weisong Shi}
\affiliation{%
  \institution{University of Delaware}
  \city{Newark}
  \state{DE}
  \country{USA}}

\author{Yanzhi Wang}
\affiliation{%
  \institution{Northeastern University}
  \city{Boston}
  \state{MA}
  \country{USA}
}

\renewcommand{\shortauthors}{C. Wu and Y. Gong, et al.}

\begin{abstract}
Object detection on the edge (Edge-OD) is in growing demand thanks to its ever-broad application prospects. However, the development of this field is rigorously restricted by the deployment dilemma of simultaneously achieving high accuracy, excellent power efficiency, and meeting strict real-time requirements.
To tackle this dilemma, we propose AyE-Edge, the first-of-this-kind development tool that explores automated algorithm-device deployment space search to realize \uline{A}ccurate \uline{y}et power-\uline{E}fficient real-time object detection on the \uline{Edge}. Through a collaborative exploration of keyframe selection, CPU-GPU configuration, and DNN pruning strategy, AyE-Edge excels in extensive real-world experiments conducted on a mobile device. The results consistently demonstrate AyE-Edge's effectiveness, realizing outstanding real-time performance, detection accuracy, and notably, a remarkable 96.7\% reduction in power consumption, compared to state-of-the-art (SOTA) competitors
\end{abstract}




\maketitle

\section{Introduction}
With exceptional accuracy, DNNs-based object detector models are widely adopted for real-time detection and analysis, spanning various applications such as autonomous driving, healthcare, and sports analytics.
However, the computation/memory-intensive nature inherent in DNN detectors poses a challenge for deployment on edge devices with limited hardware resources. This challenge is further exacerbated by the stringent requirements of processing massive amounts of video data in real-time scenarios.

Numerous techniques have been proposed to enable real-time object detection on the edge (Edge-OD), addressing either detection accuracy or time/energy efficiency \cite{ren2015faster,254354}.  
Our comprehensive exploration identifies keyframe selection strategies \cite{roychowdhury2022semi}, DNN model pruning methods \cite{10247713}, and CPU-GPU configuration \cite{kim2021ztt} as standout approaches with superior performance. 
These three techniques operate at the data layer (i.e., real-time videos), detector layer (i.e., DNN models), and hardware layer (i.e., edge devices), respectively. 
Keyframe selection aims to minimize input data volume by selecting essential frames for processing (instead of all the frames captured by cameras with high redundancy), reducing time and power costs while guaranteeing detection accuracy.
Pruning methods aim to reduce both the DNN detector model size and the number of computations, enhancing time/power efficiency but potentially hurting detection accuracy.
CPU-GPU configuration tuning allows the adjustment of CPU's and GPU's voltage/frequency (V/F) levels and the selection of CPU core clusters to balance power consumption and execution speed. 
\begin{figure*}[!hbtp]
\setlength{\abovecaptionskip}{0in}
\setlength{\belowcaptionskip}{-0.17in}
    \centering
    \includegraphics[width=6.9in]{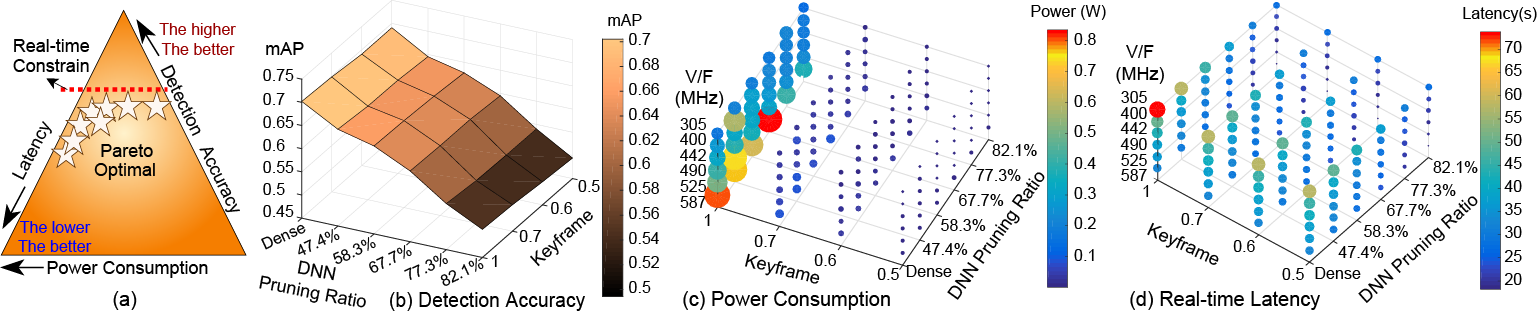}
    \caption{(a) the dilemma of achieving high accuracy, excellent power efficiency, and meeting strict real-time requirements; The impact of keyframe selection strategies, DNN model pruning methods, and DVFS techniques on (b) detection accuracy, (c) power consumption and (d) real-time performance.}
    \label{fig:motiv}
\end{figure*}
However, building a holistic framework to coordinate these techniques across different layers poses a significant challenge, as it entails navigating a complex multi-objective optimization problem with conflicting goals. Our experiments conducted on a real device shed light on the impact of these techniques on individual objectives, with detailed insights provided in Section \ref{sec:eval}. As depicted in Figure \ref{fig:motiv}, we observe that higher V/F levels or sparser DNN models can enhance real-time performance but may incur increased power consumption or a potential degradation in detection accuracy, and vice versa. Furthermore, misjudgments in keyframe selection can lead to resource wastage or accuracy degradation, underscoring the intricate nature of achieving a Pareto-optimal solution among these conflicting objectives.

To address this challenge, an exhaustive search for the best suitable combination with Pareto optimality is deemed the most intuitive and effective way. Nevertheless, this proves challenging for three main reasons. Firstly, the vast deployment space makes exhaustive search prohibitively expensive. For example, the Oneplus 8T smartphone (as detailed in Sec. \ref{sec:eval}) presents 3.67E+05 potential deployment scheme candidates for device configuration alone, and each will lead to distinct different performance-per-energy outcomes.
Secondly, the lack of a performance collector for Edge-OD deployment poses a great hurdle. It is crucial for fair comparisons of distinct deployment schemes, enabling Edge-OD developers to analyze performance and hardware costs accurately before actual implementation, in turn substantially reducing R\&D cycles. 
Lastly, another essential that has been less studied is an automated coordinator capable of intelligently searching the deployment space and generating Pareto optimal deployment schemes that meet target accuracy, power efficiency, and real-time requirements.

To this end, we propose AyE-Edge, a novel development tool designed to enable superior accurate yet power-efficient real-time Edge-OD. AyE-Edge comprises three components, including (1) an optimized Edge-OD deployment space, which features a temporal locality (T-Locality) based keyframe selector, a latency-restrained DNN pruner, and a CPU core cluster selector to optimize the vast space using a branch and bound methodology; (2) an Edge-OD performance collector that allows precise estimation of detection accuracy, power consumption, and real-time latency when deploying given DNN detectors on edge devices; and (3) a multi-agent deep reinforcement learning (MARL)-assisted coordinator, which efficiently explores and exploits the defined space, and then makes informed decisions on how to intelligently collaborate the three techniques, ensuring that all videos can be processed within time constraint while achieving the desired detection accuracy and power efficiency.
Experiments on a real mobile device (i.e., the OnePlus 8T smartphone) consistently validate AyE-Edge's effectiveness, showcasing outstanding real-time performance, detection accuracy, and a remarkable 96.7\% reduction in power consumption against the SOTAs.
To the best of our knowledge, \ul{AyE-Edge is the first framework that achieves power-efficient real-time Edge-OD with satisfactory task accuracy simultaneously with the systematic coordination among the innovations across the data layer, detector layer, and hardware layer.}

\begin{figure}[!htbp]
\setlength{\abovecaptionskip}{0in}
\setlength{\belowcaptionskip}{-0.12in}
\centering
\includegraphics[width=0.9\linewidth]{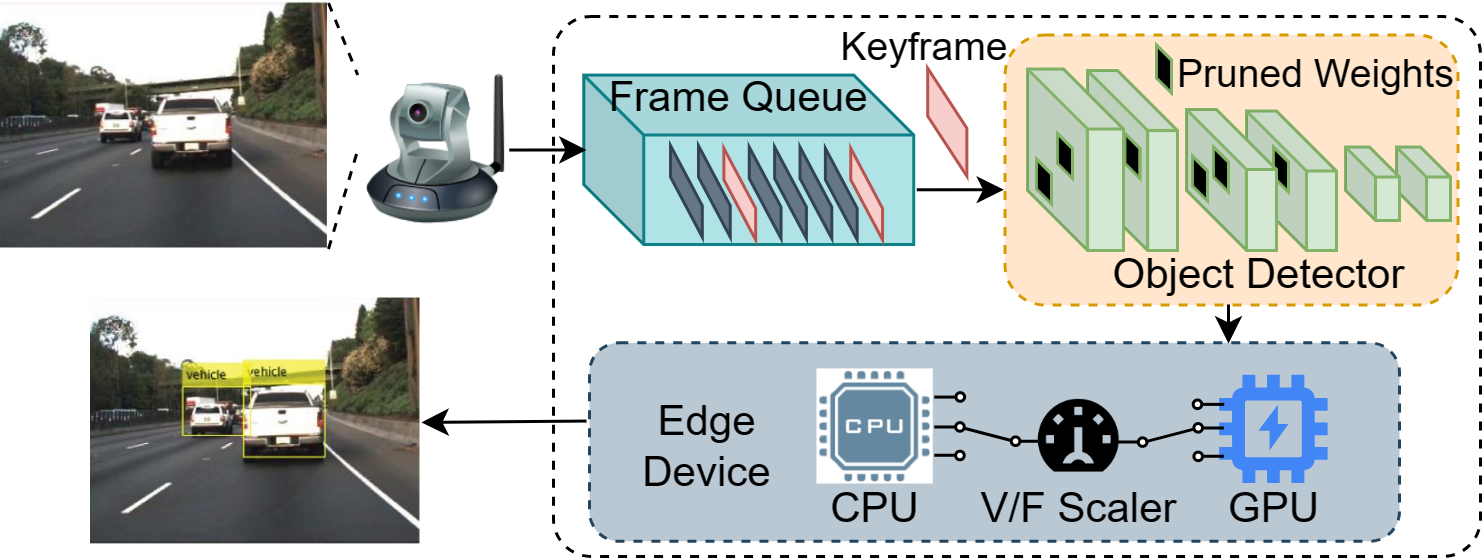}
\caption{The architecture for object detection on the edge.}
\label{fig:odedge}
\end{figure}
\section{Background and Related Work}
Fig.~\ref{fig:odedge} illustrates the architecture of an Edge-OD system, where a frame queue is managed either near the camera or by the OD application.
As frames queue up, the system selectively feeds a frame, denoted as a \texttt{keyframe}, to the detector (i.e., typically a DNN model) as the input.
The adopted DNN models (e.g., YOLO \cite{yolov5}) are often pruned for acceleration before being deployed on edge devices. These devices commonly are heterogeneous SoC platforms encompassing both CPUs and GPUs. The modern CPUs/GPUs occupy DVFS (Dynamic Voltage and Frequency Scheduling) capability (e.g., V/F scaler shown in Fig.~\ref{fig:odedge}), supporting the fine-grained adjustment of voltage/frequency levels for CPU/GPU.
Meanwhile, the current edge CPU commonly adopts ARM big.\textsc{little} micro-architectures \cite{big.little}, enabling the task scheduling among different CPU core clusters.

\textbf{Keyframe selection techniques.} They are commonly adopted to reduce input data volume in video processing tasks.
Offline tasks frequently leverage DNN-assisted methods for keyframe selection \cite{gowda2021smart, tang2023deep, badamdorj2022contrastive, jiang2022joint, narasimhan2021clip, badamdorj2021joint}, achieving high accuracy but posing intensive computational burdens.
In contrast, real-time OD predominantly adopts static threshold-based methods \cite{sara2019image, roychowdhury2022semi, wang2004image}. These methods identify the current frame as a keyframe if its feature similarity to the last keyframe is below a given threshold.
Various similarity features, such as structure similarity index measure (SSIM) \cite{sara2019image}, edge, corner \cite{li2020reducto} have been introduced.
Among them, SSIM is the most widely used feature, encapsulating luminous intensity, contrast ratio, and structure among frames.
However, the main drawback of the threshold-based selection methods is they cannot adapt to the dynamic patterns of input videos, while this ability is vital \cite{li2020reducto}.
Taking object detection in autonomous driving as an example, traffic environments change all the time, especially in urban areas. The selection of keyframes must be adaptive (rather than predefined by users) and lightweight to ensure detection accuracy.

\textbf{DNN Detector models.} They can be classified into two types: One-stage detectors and Two-stage detectors. One-stage detectors, e.g., YOLO \cite{yolov5}, SSD \cite{liu2016ssd}, and their subsequent works \cite{cai2021yolobile, wang2023yolov7}, extract bounding boxes from
input frames and regress bounding boxes for detected objects.
These algorithms are mainly optimized to balance the trade-off between model accuracy and inference speed, especially
to achieve real-time performance.
In contrast, two-stage detectors,  e.g., R-CNN \cite{girshick2014rich}, Fast R-CNN \cite{girshick2015fast}, Faster
R-CNN \cite{ren2015faster}, Mask R-CNN \cite{he2017mask}, pay more attention to task
accuracy. They extract the region of interest (ROI) from input frames in the first stage and perform classification and bounding box regression based on the ROI in the second stage.
These methods might achieve higher task accuracy but take a longer inference time, e.g., roughly 10$\times$ longer inference time in our experiments, making them less appealing for real-time Edge-OD.

\textbf{DNN pruning.}
Given the redundancy in DNN models, various pruning approaches are introduced \cite{ma2022blcr, niu2020patdnn}.
They mainly pursue an aggressive pruning ratio with minor accuracy loss.
Unstructured pruning methods prune net weights to reduce the memory overhead at arbitrary locations, which cannot accelerate DNNs because
of the irregular network sparsity \cite{10.1145/3613424.3614312}.
Structured pruning,
e.g., channel-based \cite{wen2016learning}, kernel-based \cite{zhong2021revisit}, block-based \cite{ma2022blcr}, removes the whole channel/block/CNN (convolutional neural networks) kernel to boost the
DNN inference, which might lead to a lossy accuracy. Kernel pattern-based pruning \cite{niu2020patdnn} prunes CNN kernels into specific patterns.
These methods mostly focus on chasing an aggressive pruning ratio with minor or no accuracy lost.
Recent works \cite{song2021dancing, 10247713, gong2022all, yu2018slimmable} support run-time pruning ratio reconfiguration.
Specifically, Gong {\it et al.} \cite{10247713} propose a soft mask-based pruning ratio reconfiguration approach, which supports run-time reconfiguration for pre-trained sparse DNNs by comparing the network parameter importance in the soft mask with a given threshold without further retraining.
These works provide solid technical support for automated deployment exploration of real-time Edge-OD.

\textbf{CPU-GPU Configuration.} Edge devices often employ heterogeneous SoC architecture, where CPUs and GPUs are both equipped. CPU processors mostly adopt ARM big.\textsc{little} micro-architectures. 
Different CPU core clusters provide different computation capabilities and power consumption, and could be preferred by different DNN-related tasks. This becomes a challenge for system designers.
Moreover, \textbf{DVFS scheduling} supports multiple optional operating frequencies for each processor \cite{kim2021ztt}.
A higher frequency enhances processing speed but at the expense of increased power consumption.
Although the rich frequency options offer abundant flexibility in workload adaption, selecting the best-suited frequency for each processor makes the aforementioned challenge even more intensive.

\section{The Proposed AyE-Edge Tool}
\begin{figure}[!hbtp]
\setlength{\abovecaptionskip}{0in}
\setlength{\belowcaptionskip}{-0.15in}
\centering
\includegraphics[width=1\linewidth]{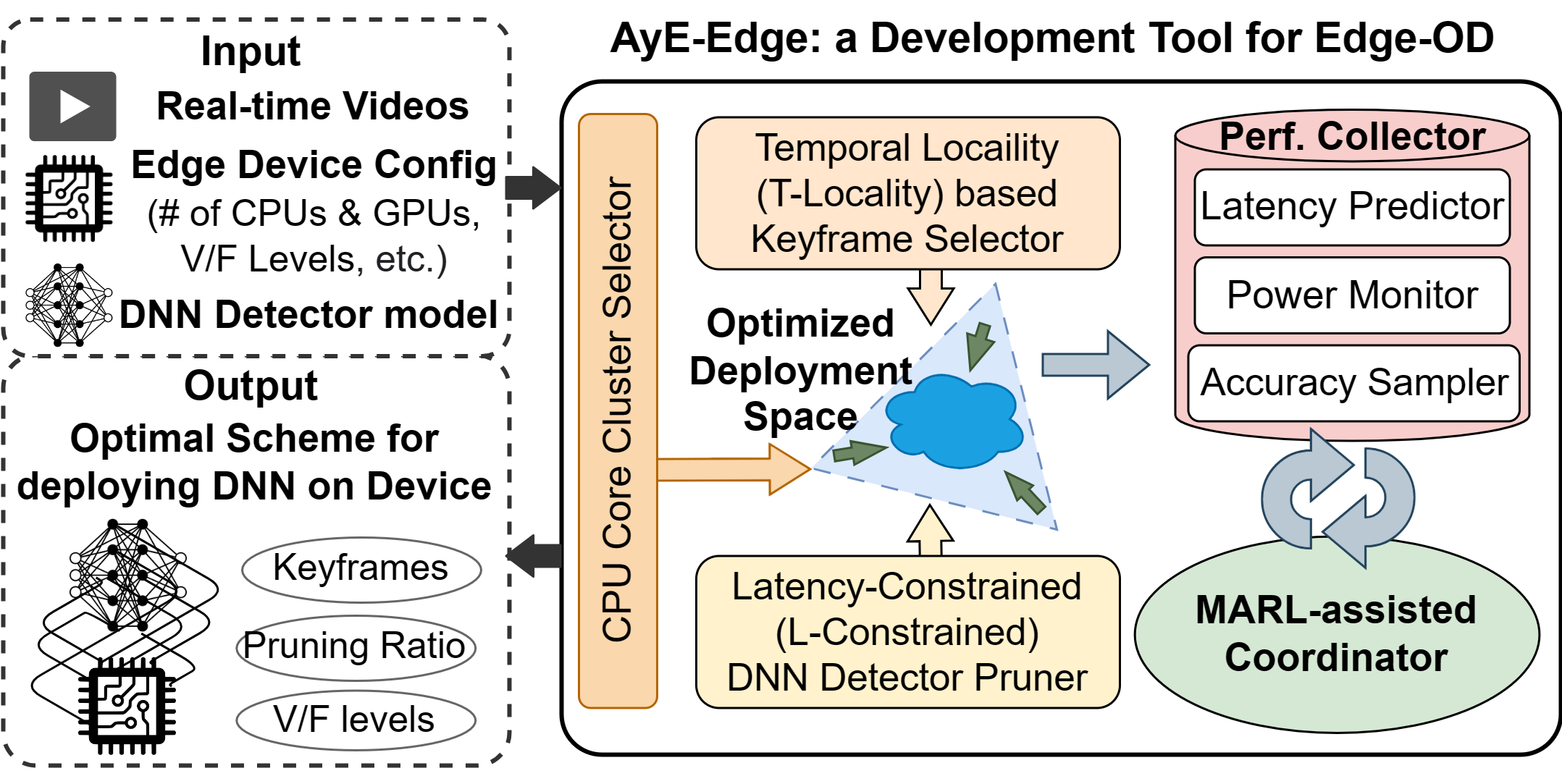}  
\caption{The proposed AyE-Edge development tool.}
\label{fig:framework}
\end{figure}

\textbf{Overview and Workflow.} As depicted in Fig.~\ref{fig:framework}, AyE-Edge comprehensively considers keyframe selection, CPU-GPU configuration, and DNN pruning for Edge-OD systems to address the aforementioned intricate trade-off dilemma involving detection accuracy, real-time speed, and power consumption.
Upon the arrival of a new frame, AyE-Edge checks whether the frame is identified as a keyframe and the frame queue is empty. For filtered frames and an empty queue, AyE-Edge remains inactive. Otherwise, AyE-Edge initiates its workflow.
During training, the \texttt{MARL-assisted coordinator} first selects an action using the $\varepsilon$-Greedy algorithm \cite{wu2020pruning}. This action decides the deployment scheme, encompassing the next keyframe, V/F levels of CPUs and GPUs, and DNN pruning ratio.
To shrink the vast deployment space, an \texttt{optimized deployment space} is constructed, in which we put forward a T-Locality-based keyframe selector and an L-Constrained DNN detector pruner to bound the space for efficient exploration.
Moreover, a \texttt{CPU core cluster selector} is introduced to select the proper CPU core cluster beforehand.
The Edge-OD system is then configured according to the selected action.
Meanwhile, the MARL model calculates the Q-value of the action by utilizing information such as streaming frames, DNN detector mode parameters, edge device configurations, and historical actions as inputs.
Once the processing for the current keyframe is finished, the reward for the last action is calculated by collecting metrics including detection accuracy, real-time speed, and power consumption during action implementation.
The \texttt{Performance Collector} serves as an accurate evaluation tool in the AyE-Edge system, using lightweight latency models, PyTorch, and the API of our adopted power monitor, to collect these metrics.
Finally, the calculated reward is fed back to the output of the MARL coordinator for back-propagation training.

\textbf{Implementation.}
AyE-Edge could be embedded as a module in the DNN compiler, e.g., CoCo-Gen \cite{liu2020cocopie}. It provides an interface for the developers of the object detection apps.
The interface is called upon the arrival of each new frame. It takes the SSIM feature of the new frame, frame queue information, DNN detector details, and edge device configurations as inputs, and alters the deployment scheme by identifying the action with the maximum Q-value.
The selected action, i.e., the selected combination of keyframe, DNN pruning ratio, and V/F levels of CPU and GPU, is returned as the output of the interface function for users to configure the Edge-OD system.

\subsection{Optimized Edge-OD Deployment Space}
The deployment space for Edge-OD consists of keyframe selection choices, DNN pruning ratios, and configurations of the CPU-GPU platform, which is a large and discrete space that is prohibitively expensive to search.
Therefore, we propose three knobs in AyE-Edge to strategically shrink the initial space: a temporal locality (T-Locality)-based keyframe selector, a latency-constrained (L-Constrained) DNN detector pruner, and a CPU core cluster selector. These knobs act on the keyframe selection, the DNN pruning processes, and the CPU-GPU heterogeneous platform configuration. According to the characteristics of fed real-time videos, they set reasonable bounds adaptively to narrow their ranges. These knobs play a pivotal role in streamlining the exploration process and facilitating the identification of Pareto-optimal deployment schemes.

\begin{figure}[!htbp]
\setlength{\abovecaptionskip}{0in}
\setlength{\belowcaptionskip}{-0.15in}
\centering
\includegraphics[width=.9\linewidth]{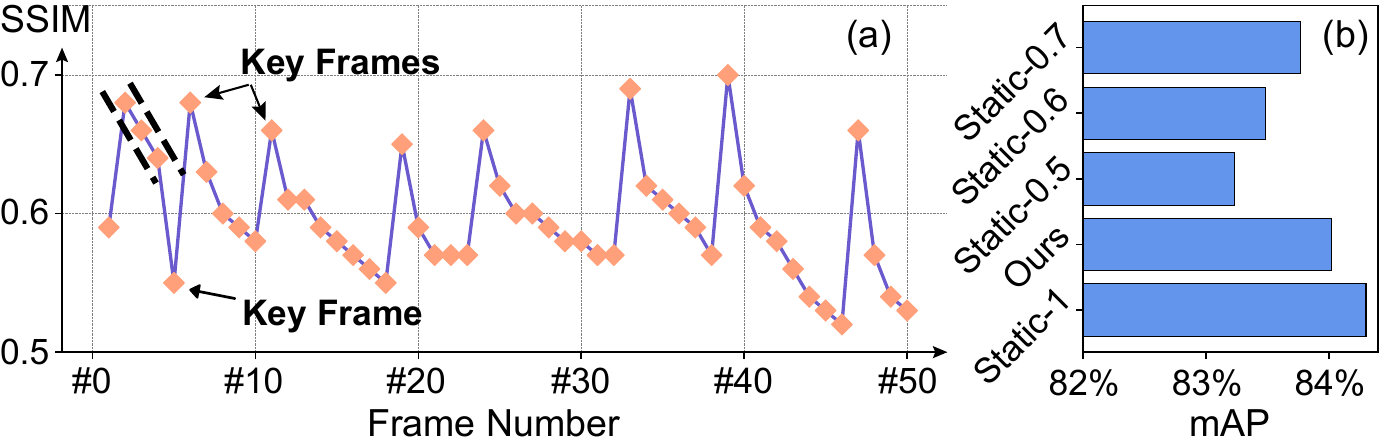}
\caption{(a) SSIM features of all the frames from a video clip based on YOLO-v5 detectors with the BDD100K dataset; (b) The mAP comparison among methods.}
\label{fig:ssim}
\end{figure}

\textbf{T-Locality-based Keyframe Selector.} This knob is designed to establish the lower bound for effective keyframe selection. It is conceptualized based on a key observation derived from extensive experimental results. That is, although frame feature similarity may exhibit randomness over the long term, it demonstrates a local regular pattern.
In Fig.~\ref{fig:ssim} (a), we visualize the SSIM features of all the frames from a short video clip. This experiment is based on the BDD100K dataset \cite{yu2020bdd100k}. The results indicate a gradual decline in feature similarities (i.e., SSIM in this figure) between the keyframe and several subsequent frames, fitting a linear regression.
Similar observations could be found on other features, e.g., edge, and corner.
Therefore, we can predict the similarity between the subsequent frames and the current keyframe by considering a range around the regressing line, such as 5\% of the similarity between the current keyframe and its next frame.
When a subsequent frame emerges whose similarity with the current keyframe exceeds this range, We define this frame as the new keyframe. That is because the new and current keyframes exhibit irregular changes in their feature similarities and should be processed individually. 

Based on such an observation, we compare the mean average precision (mAP) of object detection tasks when adopting this selection method (short for \textit{Ours}) against the SOTA static threshold-based selection methods \cite{sara2019image} given thresholds of 0.5, 0.6, and 0.7 (i.e., \textit{Static-0.5}, \textit{Static-0.6}, and \textit{Static-0.7}, respectively). \textit{Static-1} denotes the method of processing all frames in the video instead of keyframes only (i.e., without frame filtering), which achieves the highest accuracy and is referred to as the upper bound of keyframe selection.
Fig.~\ref{fig:ssim} (b) shows the accuracy comparison results, which verifies the correctness of our selector. It delivers higher accuracy over SOTAs, close to that of \textit{Static-1} with a tiny decrease of 0.28\%.

Therefore, we define the frame number obtained through our T-locality-based keyframe selector as \textit{\#Ours} and the total number of frames as \textit{\#Total}, the available range for keyframe selection spans from \textit{\#Ours} to \textit{\#Total}.
In addition, we further delve into the lower limit for keyframe selection. Beyond considering the distribution of keyframe similarity, application requirements need to be considered as well.
For instance, in autonomous driving where the safety response time is 400ms \cite{zhang2019determinants}, the driving system should at least respond within 12 frames for an RGB camera with a sampling frequency of 30ms, allowing sufficient time for frame processing. In such scenarios, the lower bound should be determined as the minimum between 12 frames and \textit{\#Ours}.

\textbf{L-Constrained DNN Detector Pruner.}
To further narrow down the search space in AyE-Edge, we establish upper and lower bounds for pruning ratios.
The lower bound is determined by the highest pruning ratio which does not result in any loss of model accuracy. This value is often empirically derived from extensive experimental results, utilizing the accuracy plug-in that we develop for Edge-OD, as elaborated in Sec.~\ref{subsec:perf}.
The upper bound is user-defined. One approach is to define the upper bound as the pruning ratio at which the model accuracy experiences an acceptable reduction, such as 5\%. Alternatively, users may choose to set a more aggressive upper bound, considering the pruning ratio at which the DNN model fails to converge if exceeded.

With the lower and upper bounds for the DNN pruning ratio established, we can leverage the DNN pruning technique proposed in \cite{10247713} to implement run-time reconfiguration of the pruning ratios for DNN detectors in AyE-Edge.

\textbf{CPU Core Cluster Selector.}
One interesting observation from our experiments is that, although most DNN tasks have been offloaded to GPU, CPU is a main contributor to device power consumption \cite{halpern2016mobile}.
By simply scheduling DNN-involved tasks to the `little' core cluster, the power consumption of DNN inference could be decreased by roughly 60\%, with harmless inference speed.
This observation motivates us to integrate CPU core cluster selection into our AyE-Edge.

Through investigation, we found that different clusters impact the power efficiency of DNN inference mainly by different cache sizes if in the same CPU V/F level.
Given CPU mainly takes charge of data transferring in DNN inference \cite{wu2023iccad}, the on-chip CPU cache size decides the data exchange times between CPU and DRAM, while data transmission speed is influenced by the V/F level of the core cluster.
Thus, the activation of the CPU core cluster could be selected by comparing the cache size of different clusters with the DNN layer which is the largest in size of weights and output.
With this heuristic method, AyE-Edge could select the proper CPU core cluster before DVFS scaling, decreasing the deployment space of embedding the CPU core selection in the MARL-assisted Coordinator.

\subsection{Edge-OD Performance Collector} \label{subsec:perf}
Our Performance Collector is designed to precisely collect or estimate the detection accuracy, latency, and power consumption for processing keyframes. These pieces of information are prerequisites for the training of the MARL-assisted coordinator. It comprises three integral sub-assemblies:
\begin{enumerate}[(1)]
    \item a Power Monitor \cite{Moonson} to measure precisely how much power the edge devices are drawing for processing one keyframe.
    \item an Accuracy Sampler to emulate the detection accuracy according to the given real-time keyframes, DNN models with predefined pruning ratio, and hardware devices with preconfigured V/F levels. This Sampler is developed as a lightweight plug-in utilizing an open-source machine-learning framework (e.g., PyTorch), as does in \cite{li2023automated}.
    \item a Latency Predictor to accurately forecast the processing time of keyframes with negligible extra overhead.
\end{enumerate}

Due to space limitations, we focus on detailing the proposed Latency Predictor in this section. Both the Power Monitor and the Accuracy Sampler have been well-studied in prior arts \cite{Moonson, li2023automated}.
Based on the given CPU-GPU platforms with DVFS capability, the processing time for a keyframe can be formulated as (\ref{eq:dvfs_pred}).
\begin{align}
\label{eq:dvfs_pred}
L_{pred.} = \mathop{\underbrace{L_{dense} \times \frac{VF^G_{max}}{VF^G_{cur}}}}\limits_{GPU~V/F~level~pred.} + \mathop{\underbrace{\gamma (\frac{VF^C_{max}}{VF^C_{cur}} - 1)}}\limits_{CPU~V/F~level~pred.}
\end{align}
where $L_{pred.}$ is the predicted latency with the given CPU V/F level (i.e., $VF^C_{Cur}$ ) and GPU V/F level (i.e., $VF^G_{cur}$ ); $L_{dense}$ is the dense DNN model speed with the selected CPU core cluster and the highest V/F levels; $VF^C_{max}$ and $VF^G_{max}$ are the highest V/F level for CPUs and GPUs, respectively; $\gamma$ is a parameter related to the largest DNN layer size and CPU cache size, which is modeled as $\gamma=\frac{max(\textsc{MEM}_{\ell i})}{\textsc{Cache}_{CPU}}$. $\textsc{MEM}_{\ell i}$ is the memory footprint of DNN layer $\ell i$ (including weights and output feature-map) and $\textsc{Cache}_{CPU}$ is the total L2-cache size of selected CPU cluster.

The formulation of the latency model is guided by three key insights. Firstly, both CPUs and GPUs play important roles in contributing to DNN inference latency, exhibiting a linear relationship with their respective frequency \cite{tang2019impact, wu2023iccad}.
Secondly, differing from GPUs which are mainly in charge of MAC calculations, CPUs are primarily responsible for managing the pre/post-calculation stages in DNN inference \cite{wu2023iccad}. Due to the sequential interdependence between MAC calculation procedures and pre/post-calculation stages, the contributions of CPUs and GPUs to DNN latency are additive.
Thirdly, at the same V/F level, the impact of CPUs with different configurations on DNN latency is predominantly determined by their cache sizes and maximum memory bandwidth, thus we model the coefficient $\gamma$ as such.

Furthermore, the adoption of DNN pruning techniques complicates Edge-OD latency prediction. Unlike DVFS levels, the relationship between the DNN pruning ratio and inference speed is more intricate and cannot be straightforwardly approximated \cite{yuan2021mest}.
We propose to construct the prediction model using a look-up table (LUT) generated during the training stage of the DNN pruning algorithm. This table records the latencies of the given DNN detectors under predetermined pruning ratios when deployed on edge devices with designated hardware configurations. We extend (\ref{eq:dvfs_pred}) to Eq.~\ref{eq:prune}, where $L_{pr=i}$ is the latency of the DNN model under the pruning ratio of $i$, which can be easily acquired by table-checking.
It is noteworthy that the look-up table does not necessarily bond to the DNN training process.
Instead, it can be manually created through evaluations based on user-demanding pruning ratios.
\begin{align}
\label{eq:prune}
L_{pred.} = L_{pr=i} \times \frac{VF^G_{max}}{VF^G_{cur}} + \gamma (\frac{VF^C_{max}}{VF^C_{cur}} - 1) 
\end{align}

\begin{figure}[!htbp]
\centering
\setlength{\abovecaptionskip}{0in}
\setlength{\belowcaptionskip}{-0.2in}
\includegraphics[width=1\columnwidth]{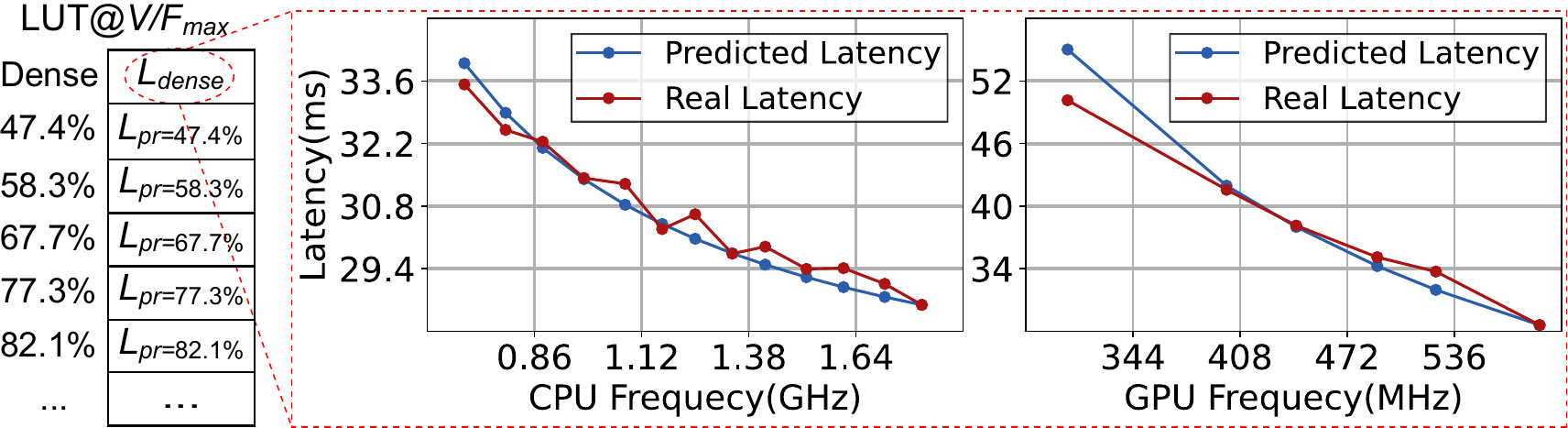}
\caption{The predicted latencies of our Latency Predictor in Eq.~(\ref{eq:prune}) across (a) various CPU V/F levels with GPU fixed with the highest GPU frequency and (b) various GPU V/F levels with CPUs fixed with the highest frequency. Assumed $VF^C_{max}=1.8GHz$, $VF^G_{max}=587MHz$, and $\gamma=3.36$.}
\label{fig:modelvalid}
\end{figure}

\uline{Model Validation.} We verify the precision of the proposed Latency Predictor in (\ref{eq:prune}) based on the YOLO-v5 detector and the BDD100K dataset. As depicted in Fig.~\ref{fig:modelvalid}, our Predictor achieves high prediction accuracy, with only a 1.9\% average error (3.6\% at maximum) across different CPU V/F levels and a 3.1\% average error (8.1\% at maximum) across all GPU V/F levels.

\subsection{MARL-assisted Coordinator}
\label{subsec:marl}
\begin{figure}[!hbtp]
\setlength{\abovecaptionskip}{0in}
\setlength{\belowcaptionskip}{-0.15in}
\centering
\includegraphics[width=0.6\linewidth]{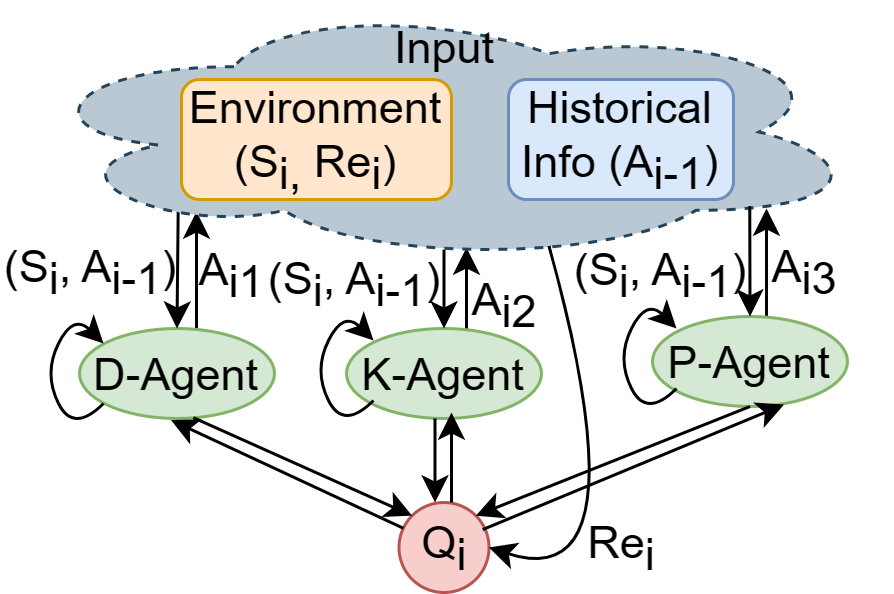}  
\caption{MARL-assisted coordinator.}
\label{fig:marl}  
\end{figure}

To find the Pareto-optimal front in the optimized deployment space of AyE-Edge, we adopt a reinforcement learning (RL) model.
However, the model is hard to converge owning to the overlarge action space.
In Aye-Edge, the action pool of the RL model consists of all combinations of candidate keyframes, pruning ratios, and V/F levels of CPU and GPU, which leads to over 10,000 actions in total.
To address this issue, we adopt a multi-agent  RL model (MARL), as shown in Fig.~\ref{fig:marl}.
MARL consists of three parallel agents, including \texttt{D-Agent} (DVFS agent), \texttt{K-Agent} (Keyframe agent), and \texttt{P-Agent} (Pruning ratio agent).
All agents perceive the same environment (state $S_i$, reward $Re_i$), and hence the same Q value in each iteration.
Furthermore, each agent consists of an RNN network to take the same Q value, together with the last actions of all agents ($A_{i-1}$) of the last iteration as the input.
In this manner, all agents communicate with each other to work collaboratively.

\textbf{State space.}
The state space in the Edge-OD environment includes the status of the frame queue, DNN object detector, and edge devices.
Specifically, it includes 1) SSIM of the current frame, pending frame number in the queue, and the current keyframe; 2) DNN parameter, including network weight number, channel number, layer number, CNN kernel number and size in each layer, and the width of each fully connected layer; 3) The configuration of the edge device, including the current CPU and GPU frequency, the maximum memory bandwidth, and the on-chip cache size of the CPU and GPU.

\textbf{Action space.}
The agents in MARL have different action spaces.
Specifically, the action space of D-Agent, K-Agent, and P-Agent includes all V/F levels of CPU and GPU, candidate keyframes to select, and all pruning ratios, respectively.
The outputs of RNN in each agent are the Q-values of all actions for each agent.
\begin{equation}
\label{eq: performance ceiling roofline}
  Reward = 
    \begin{cases}
       Acc_{N} - \alpha\times Po_{N} \ \ \ \ \ \ \ \ \ \ \ \ s.t.~L_{pred.} \leq RT_{tar}
      \\
        Acc_{N} - \alpha\times Po_{N} - SP \ \ \ s.t.~ L_{pred.} \textgreater RT_{tar}
    \end{cases}
\end{equation}

\textbf{Reward.}
The real-time constraints in Edge-OD could be classified as soft and hard constraints.
We formulate the reward in our deep Q-learning model as described in Eq.~\ref{eq: performance ceiling roofline}, where $SP$ is a soft penalty calculated by $(L_{pred.} - RT_{Tar})/RT_{Tar}$.
Users could replace $SP$ with other penalties as needed, e.g., a constant value as a hard penalty for hard real-time constraint scenarios.
$Acc_N$ and $Po_N$ are the mAP and average power consumption of detected keyframes in each video, respectively.
$L_{pred.}$ is the predicted latency of the processed keyframe during the implementation of the corresponding action.
$RT_{tar}$ is the real-time constraint, e.g., 33ms.
$\alpha$ is a factor assigned by users to select an accuracy- or power-oriented tuning, which is 1 by default.
Note that, the average power consumption is calculated by the power consumption of all keyframes divided by that of all frames in the current video.
For filtered frames when there are frames in the queue, AyE-Edge will still be activated and alter the deployment scheme of the Edge-OD system accordingly, whose reward is calculated by assigning $Acc_N$ in Eq.~\ref{eq: performance ceiling roofline} as 0.

\textbf{Model Training.}
We adopt a two-layer RNN (multi-layer perceptron) as the Deep-Q-Network (DQN) in each agent of MARL, taking the factors of the state space and the last action of all agents as the input, and the Q-value of different agent actions as the output.
MARL is trained with three real devices, including Samsung S20, Jetson Board, and Google Pixel 6.
Note that, the MARL model needs to be fine-tuned in the initial stage of deployment in each edge device.
During initialization, AyE-Edge will take several frames from the camera and feed the device information, DNN objector parameter, and frame queue status to the MARL model for fine-tuning, so that AyE-Edge can adapt to different hardware platforms and DNN models. 


\section{Performance Evaluation}
\label{sec:eval}
\textbf{Experimental Platform.} We adopt OnePlus 8T mobile phone as our experimental platform, which is equipped with Qualcomm SnapDragon 865 chipset with a Qualcomm Kryo 585 Octa-core CPU (1$\times$2.84 GHz Cortex-A77 \& 3$\times$2.42 GHz Cortex-A77 \& 4$\times$1.80 GHz Cortex-A55) and a Qualcomm Adreno 650 GPU. 
Table~\ref{tab:possible actions} shows the DVFS levels for CPU and GPU on the mobile phone.
The phone is rooted to tune the CPU and GPU frequency levels.
The power consumption is measured by Monsoon High Voltage Power Monitor (HVPM)during the DNN inference \cite{Moonson}. 
CoCo-Gen~\cite{liu2020cocopie} compiler is adopted to accelerate the inference on mobile devices.
Each test takes 100 runs on different configurations for a DNN. 
We take $\Delta P=P_{avg}-P_{idle}$ as the power consumption for the DNN inference, where the $P_{avg}$ is the average power during DNN inference and $P_{idle}$ is the power of the idle state.
The idle state is defined as the minimum CPU and GPU frequency, and unnecessary functionalities and applications are disabled, such as Wi-Fi, Bluetooth, and Screen. To meet real-time constraints, which should be more than 30 frames per second,  $RT\_{tar}$ in Sec.~\ref{subsec:marl} is set as 33ms.
\renewcommand{\arraystretch}{1.1}
\begin{table}[!htbp]
\centering
\begin{tabular}{@{}ccc@{}}
\toprule
\begin{tabular}[c]{@{}c@{}}CPU Cluster\\ (\#Cores)\end{tabular} & \begin{tabular}[c]{@{}c@{}}$f_{CPU}$\\ (GHz)\end{tabular}                                                                                                                                & \begin{tabular}[c]{@{}c@{}}$f_{GPU}$\\ (MHz)\end{tabular}                                    \\ \midrule
\multicolumn{1}{c|}{little (4)}                                 & \multicolumn{1}{c|}{\begin{tabular}[c]{@{}c@{}}0.69, 0.78, 0.88, 0.97, 1.08, 1.17, \\ 1.25,  1.34, 1.42, 1.52, 1.61, 1.71,  1.80\end{tabular}}                                           & \multirow{7}{*}{\begin{tabular}[c]{@{}c@{}}587\\ 525\\ 490\\ 441.6\\ 400\\ 305\end{tabular}} \\ \cline{1-2}
\multicolumn{1}{c|}{medium (3)}                                 & \multicolumn{1}{c|}{\begin{tabular}[c]{@{}c@{}}0.71, 0.83, 0.94, 1.06, 1.17, 1.29, \\ 1.38, 1.48, 1.57, 1.67, 1.77, 1.86, \\ 1.96, 2.05, 2.15, 2.25, 2.34, 2.42\end{tabular}}            &                                                                                              \\ \cline{1-2}
\multicolumn{1}{c|}{big (1)}                                    & \multicolumn{1}{c|}{\begin{tabular}[c]{@{}c@{}}0.84, 0.96, 1.08, 1.19, 1.31, 1.4, 1.52, \\ 1.63, 1.75, 1.86, 1.98, 2.07, 2.17, 2.27, \\ 2.36, 2.46, 2.55, 2.65, 2.75, 2.84\end{tabular}} & \\ \bottomrule
\end{tabular}
\caption{DVFS levels for CPU and GPU on OnePlus 8T. The number of CPU cores in the cluster is shown after the cluster name.}
\label{tab:possible actions}
\end{table}

\begin{table}[!htbp]
\centering
\begin{tabular}{ccccc}
\toprule
\# of objects            & 0-10  & 10-20       & 20-30 & \textgreater 30  \\ \midrule
\# of videos        & 104& 85 &10&1  \\\bottomrule
\end{tabular}
\caption{Video object number distribution.}
\label{table: video proposal}
\end{table}

\textbf{Benchmarks and Datasets.} We adopt two commonly used one-stage object detectors in our experiments to evaluate the effect of AyE-Edge.
The first detector is YOLO-v5, which is lightweight and can achieve real-time DNN inference without pruning.
The second detector SSD, has to be pruned to meet the real-time constraint.
We evaluate the task performance and power consumption using the BDD100K dataset \cite{yu2020bdd100k} with 200 sampled videos. Each video maintains a uniform length of 40 seconds, with a resolution of 720p and a frame rate of 30 FPS. The object number distribution of these videos is shown in Table~\ref{table: video proposal}.

\begin{table*}[h!]
\centering
\begin{tabular}{c|c|c|c|c|c|c|c|c|c}
\toprule
 Model & Approach & Feature & Method & L/F &KF \# & WT  &WP& P/V & mAP\\ \midrule
\multirow{7}{*}{\rotatebox{90}{YOLO}} & Origin & N/A&  N/A & 26.1 & 1174 & 0.0 & 0.0\% & 3.04 &  70.2 \\
& ST+AIO      &   SSIM  & Thre-based (0.7)  & 26.1 & 125.9 & 0.0 & 0.0\% & 0.33 & 68.6   \\
& ST+Herti    &   SSIM  & Thre-based (0.7) & 28.7  & 184.8 & 0.0 & 0.0\% & 0.15 & 69.9       \\
& Reducto+AIO &   Edge  & K-means & 28.7 & 111.6 & 0.0 & 0.0\%  & 0.29 & 68.9     \\
& Reducto+Herti & Edge  & K-means & 28.7 & 178.2 & 0.0 & 0.0\% & 0.15 & 68.9     \\
& ST+AIO-L & SSIM  & Thre-based (0.7) & 24.3 & 125.9 & 0.0 & 0.0\% & 0.02 & 51.1    \\ 
& \cellcolor[gray]{.9}AyE-Edge-C & \cellcolor[gray]{.9}SSIM  & \cellcolor[gray]{.9}Dynamic &
\cellcolor[gray]{.9}32.2 & \cellcolor[gray]{.9}171.5 & \cellcolor[gray]{.9}0.0 & \cellcolor[gray]{.9}0.0\% & \cellcolor[gray]{.9}0.10 & \cellcolor[gray]{.9}70.3    \\
& ST+AIO-L & SSIM  & Thre-based (0.7) & 24.3 & 125.9 & 0.0 & 0.0\% & 0.02 & 51.1    \\ 
& \cellcolor[gray]{.9}AyE-Edge-L & \cellcolor[gray]{.9}SSIM  & \cellcolor[gray]{.9}Dynamic &
\cellcolor[gray]{.9}34.1 & \cellcolor[gray]{.9}152.9 & \cellcolor[gray]{.9}0.1 & \cellcolor[gray]{.9}2.2\% & \cellcolor[gray]{.9}0.01 & \cellcolor[gray]{.9}64.0    \\      
\midrule
\multirow{8}{*}{\rotatebox{90}{SSD}} & Origin & N/A & N/A & 62.5 & 1174 & 28.5 & 97.8\% & 4.82 &  52.9 \\
& ST+AIO & SSIM & Thre-based (0.7)  & 30.5  &125.9 & 0.0 &0.0\% & 0.43 & 36.4   \\ 
& ST+Herti & SSIM & Thre-based (0.7) & 63.2 & 184.8 & 4.6 & 15.4\% & 0.56 & 52.4  \\ 
& Reducto+AIO & Edge   & K-means & 30.5 & 111.6 & 0.0 & 0.0\% & 0.39 & 36.4    \\ 
& Reducto+Herti & Edge & K-means & 63.2 & 178.2 & 4.43 & 14.9\% & 0.54 & 52.4     \\
& \cellcolor[gray]{.9}AyE-Edge-C & \cellcolor[gray]{.9}SSIM  & \cellcolor[gray]{.9}Dynamic &
\cellcolor[gray]{.9}34.8 & \cellcolor[gray]{.9}111.5 & \cellcolor[gray]{.9}0.2 & \cellcolor[gray]{.9}5.5\% & \cellcolor[gray]{.9}0.27 & \cellcolor[gray]{.9}50.0    \\ 
& ST+AIO-L & SSIM   & Thre-based (0.7) & 50.9 & 125.9 & 1.8 & 10.5\% & 0.07 & 36.4    \\ 
& \cellcolor[gray]{.9}AyE-Edge-L & \cellcolor[gray]{.9}SSIM  & \cellcolor[gray]{.9}Dynamic &
\cellcolor[gray]{.9}45.1 & \cellcolor[gray]{.9}83.2 & \cellcolor[gray]{.9}1.0 & \cellcolor[gray]{.9}6.9\% & \cellcolor[gray]{.9}0.08 & \cellcolor[gray]{.9}41.5    \\ 
\bottomrule
\end{tabular}
\caption{Performance comparisons of AyE-Edge and baseline approaches. \textbf{Feature} is the image feature adopted in keyframe selection. \textbf{Method} is the method of keyframe selection, and \textbf{Dynamic} is the dynamic keyframe selection method in AyE-Edge. \textbf{L/F} is the average latency per frame in milliseconds.  \textbf{KF \#} is the average keyframe number detected by the keyframe selection of approaches. \textbf{WT} is the average waiting time of blocked keyframes in milliseconds. \textbf{WP} is the percentage of blocked keyframes in all. \textbf{P/V} is the average power for each video in W. \textbf{Thred-based} in \textbf{Method} is the similarity threshold-based frame filtering method in keyframe selection.  
}
\label{table: DNN models A&M intensity}
\end{table*}

\textbf{Baselines.} We compare our work with multiple SOTAs.
The first one is \textbf{Origin}, which performs object detection on all frames of input videos without filtering, under the highest CPU and GPU DVFS level, and without DNN pruning.
The mean average accuracy of \textbf{Origin} is deemed as the upper bound, while the power consumption of it is deemed as the lower bound.
For the keyframe selection method, many works focus on the DNN-assisted approaches \cite{tang2023deep, badamdorj2022contrastive, jiang2022joint, badamdorj2021joint}, which pose an intensive computation burden on edge object detection tasks, thus is inappropriate.
Hence, we compare AyE-Edge with the classic approach, static SSIM threshold-based approach (\textbf{ST}) \cite{sara2019image}, and a SOTA approach \textbf{Reducto} \cite{li2020reducto}.
\textbf{Reducto} divides real-time video into segments (each lasts several seconds) and clusters the segments to find the best-suited threshold for each segment, to filter redundant frames in the segment.
We evaluated the edge image feature for \textbf{Reducto}.
The segment length in \textbf{Reducto} is assigned as 1s.
For \textbf{ST}, we evaluate threshold 0.5 and 0.6 as both deliver a good task precision as shown in Figure~\ref{fig:ssim} (b).
For DVFS and DNN pruning, we evaluate All-In-One (\textbf{AIO}) \cite{gong2022all} which prunes DNNs to adapt to DVFS levels, and \textbf{Herti} \cite{han2021herti} which tunes DVFS levels to adapt to DNNs.
For a fair comparison, we tune the DVFS levels with the \texttt{Schedutil} \cite{wang2021asymo} DVFS governor for \textbf{AIO} and switch DNN pruning ratios accordingly.
We combine the SOTA keyframe selection techniques with \textbf{AIO} and \textbf{Herti} to show that SOTA works cannot balance the trade-off between task accuracy, real-time speed, and power consumption if not working collaboratively.

We evaluate AyE-Edge in both common cases (\textbf{AyE-Edge-C}) and low DVFS case (\textbf{AyE-Edge-L}). 
The latter one is frequently adopted in extreme environments, e.g., high device temperatures.
The goal is to show the powerful adaptability of AyE-Edge even when hard constraints are imposed on parts of the involved factors.

\textbf{Performance Comparison.} We show the performance comparison between Aye-Edge and other baselines in Table. \ref{table: DNN models A&M intensity}. From the results, we can observe that
AyE-Edge could deliver real-time performance with significantly reduced power consumption and excellent task accuracy. The rationale behind this is that AyE-Edge strives to achieve reduced power consumption and high task accuracy with the prerequisite of real-time speed for Edge-OD.
Note that, we pursue the Pareto optimality of the aforementioned three metrics instead of the optimality of a single metric.
For example, a processed OD task with lower power consumption and real-time speed is better than the task with faster speed but higher power consumption in common cases for Edge-OD.
More specifically, \textbf{AyE-Edge-C} does not perform DNN pruning and dynamic keyframe selection in common cases as YOLO-v5 could deliver real-time speed without pruning.
For YOLO-v5, AyE-Edge-C could achieve the highest mAP compared with other approaches, owing to the locality-based keyframe selection method.
Meanwhile, AyE-Edge-C could reduce the power consumption by 96.7\% compared with \textbf{Origin}, thanks to the best-suited combination of keyframe selection, DVFS, and DNN pruning configurations.
Compared with \textbf{ST+Herti} which also shows good power efficiency, AyE-Edge-C reduces the power consumption by 33.3\%.
The reason is two-fold.
First, Herti only selects a proper DVFS level for the YOLO model, neglecting the scheduling among CPU core clusters, which is addressed by AyE-Edge.
Second, the dynamic keyframe selection method in AyE-Edge filters more frames compared with ST and generates frames with higher quality, which leads to less power consumption and higher mAP.
We notice that the sub-optimal power consumption of \textbf{ST+AIO} is 70.0\% higher than AyE-Edge-C.
While the ST framework might be capable of filtering more frames than AyE-Edge, it relies on the \texttt{schedutil} DVFS scheduler which lacks adaptability to environmental changes and does not account for task scheduling among CPU core clusters. This limitation results in higher power consumption.

For SSD, \textbf{AyE-Edge-C} also shows excellent power efficiency.
In common cases, \textbf{ST+AIO}, \textbf{Reducto+AIO}, and AyE-Edge-C could deliver real-time performance or even close.
However, the mAP of both \textbf{ST+AIO} and \textbf{Reducto+AIO} drops significantly.
The reason is that both approaches have to rely on aggressive DNN pruning to achieve real-time latency, which could significantly hurt the task accuracy.
Moreover, AyE-Edge-C delivers the lowest power consumption among all approaches in common cases.
Compared with the Origin, AyE-Edge-C can achieve a similar mAP with 94.4\% reduced power consumption.
The reason is three-fold.
First, the dynamic keyframe selection method in AyE-Edge generates the fewest keyframes with the highest quality among all approaches, which helps to deliver the task accuracy approaches to the upper bound (Origin).
Second, AyE-Edge considers task-adaptive scheduling among CPU core clusters to reduce power consumption.
Finally, AyE-Edge configures keyframe selection, DVFS level, and DNN pruning ratio collaboratively, while other approaches cannot collaborate all these factors.
To sum up, AyE-Edge manages to balance the trade-offs between real-time performance, power consumption, and task accuracy for object detection tasks by smart coordination of involved factors.

In addition, AyE-Edge shows strong adaptability even when one of the related factors is imposed by some hard constraint.
We take the low-DVFS case as an example here, which is frequently adopted by the system in extreme cases, e.g., high device temperature.
For YOLO-v5 in low DVFS cases, both \textbf{AyE-Edge-L} and \textbf{ST+AIO-L} could deliver a real-time latency.
However, ST+AIO-L achieves this mainly by aggressive DNN pruning, which leads to lower task accuracy.
As shown in Table~\ref{table: DNN models A&M intensity}, the mAP of ST+AIO-L is 20.2\% lower than AyE-Edge-L.
For SSD in low DVFS cases, AyE-Edge-L shows better task accuracy with similar power consumption compared to ST+AIO, and AyE-Edge-L can achieve near real-time performance while ST+AIO cannot.
In AyE-Edge-L, only 6.9\% of keyframes are blocked, and the average waiting time in the queue is 1.0ms.
This is because the dynamic keyframe selection in AyE-Edge could adaptively filter more frames in this scenario.
The average processed keyframes in AyE-Edge-L is only 83.2 on SSD, and it is significantly reduced compared to ST+AIO-L.
The hint behind this is that aggressive pruning for SSD as adopted by ST+AIO-L cannot balance the trade-off between task accuracy and real-time performance well, which should be achieved by the coordination among keyframe selection, DVFS, and DNN pruning as in AyE-Edge. 
To sum up, AyE-Edge shows strong adaptability not only to the frame patterns and device run-time status but also to specific circumstances when parts of the involved factors are imposed by hard constraints.

\begin{table}[!htbp]
\centering
\begin{tabular}{c|c|c|c|c|c|c}
\toprule
 Approach &  L/F &KF \# & WT  &WP& P/V & mAP\\ \midrule
DKS &  64.1 & 111.9 & 2.9 & 9.3\% & 0.31 &  52.9 \\
DP    &  32.5 & 1174 & 0.0 & 0.0\% & 1.14 & 38.0   \\
DP-C  & 29.4  & 1175 & 0.0 & 0.0\% & 2.70 & 38.0    \\
\bottomrule
\end{tabular}
\caption{Ablation study on SSD detector. \textbf{DKS} is the dynamic keyframe selection method. \textbf{DP} is the DNN pruning and DVFS scheduling methods, and \textbf{DP-C} is the DNN pruning and DVFS scheduling methods without CPU core scheduling. The define of $L/F$, $KF \#$, $WT$, $WP$, and $P/V$ could be referred to in Table~\ref{table: DNN models A&M intensity}.
}
\label{table: ablation}
\end{table}
\textbf{Ablation Study.} We study the effect of dynamic keyframe selection (\textbf{DKS}) and DNN pruning in AyE-Edge independently on the SSD detector in the ablation study.
For DKS, we explore its variation by changing the keyframe selection frequency according to the keyframe number in the queue.
For DNN pruning, we explore its variation by changing the DVFS level and DNN pruning ratio according to the keyframe number in the queue, with (\textbf{DP}) and without CPU core scheduling (\textbf{DP-C}).
Table~\ref{table: ablation} shows the evaluation results.

The first observation is that every single component of AyE-Edge could not achieve a balanced trade-off between real-time performance, task accuracy, and power efficiency.
DKS fails to meet the real-time constraint since the inference speed is not accelerated with the higher DVFS level and the sparse DNN model.
DP fails to deliver satisfactory power consumption and task accuracy.
Both DNN pruning and keyframe selection could impact the task accuracy, hence both factors should be tuned collaboratively.
Moreover, keyframe selection could significantly reduce power consumption as the number of frames is decreased.

The second observation is that CPU core scheduling in AyE-Edge could significantly impact power efficiency.
As shown in Table~\ref{table: ablation}, DP-C takes 57.8\% more average power consumption than DP.
The prior art has indicated that CPU could significantly impact the power efficiency \cite{wu2022compiler}.
Hence, the power consumption of detectors could be effectively reduced by scheduling DNN-related tasks to the proper CPU core cluster.

\textbf{Overhead Analysis.} The timing overhead of AyE-Edge is 313us for the MARL model with pruned RNN networks \cite{ma2022blcr} and the latency predictor.
The storage overhead of AyE-Edge is incurred by the look-up table, which depends on the entry number $\mathbf{K}$ of maintained latencies of DNN models with various pruning ratios.
When $\mathbf{K}$ is set to 10, the table size is only 40 bytes, which is marginal.
In summary, the overhead of our AyE-Edge is minimal and has little to no impact on object detection tasks.

\section{Conclusion}
We proposed AyE-Edge, an innovative development tool for Edge-OD deployment.
Through a synergistic arrangement of the techniques for keyframe selection, CPU-GPU parameter configuration, and DNN pruning.
AyE-Edge enables to adaptively identify the best-suited deployment scheme according to dynamic frame patterns and runtime device statuses. 
Experimental results demonstrate AyE-Edge's capability to achieve a remarkable power consumption reduction of up to 96.7\% while delivering excellent accuracy and real-time performance.
We hope this work could provide hints for subsequent research works on real-time object detection.
In future work, we will investigate the power efficiency of multi-tenant DNNs, e.g., the perception system in autonomous vehicles, which contains multiple real-time tasks such as object detection, lane detection, and semantic segmentation.

\section*{Acknowledgement}
The research reported here was funded in whole/part by the Army Research Office/Army Research Laboratory via grant W911-NF-20-1-0167 to Northeastern University. Any errors and opinions are not those of the Army Research Office or Department of Defense and are attributable solely to the author(s). This research is also partially supported by National Natural Science Foundation of China (Grant No. 62106146 and 62202159), NSF CCF-1937500, and CNS-1909172.
\bibliographystyle{ACM-Reference-Format}
\bibliography{sample-base}


\begin{thebibliography}{43}


\ifx \showCODEN    \undefined \def \showCODEN     #1{\unskip}     \fi
\ifx \showDOI      \undefined \def \showDOI       #1{#1}\fi
\ifx \showISBNx    \undefined \def \showISBNx     #1{\unskip}     \fi
\ifx \showISBNxiii \undefined \def \showISBNxiii  #1{\unskip}     \fi
\ifx \showISSN     \undefined \def \showISSN      #1{\unskip}     \fi
\ifx \showLCCN     \undefined \def \showLCCN      #1{\unskip}     \fi
\ifx \shownote     \undefined \def \shownote      #1{#1}          \fi
\ifx \showarticletitle \undefined \def \showarticletitle #1{#1}   \fi
\ifx \showURL      \undefined \def \showURL       {\relax}        \fi
\providecommand\bibfield[2]{#2}
\providecommand\bibinfo[2]{#2}
\providecommand\natexlab[1]{#1}
\providecommand\showeprint[2][]{arXiv:#2}

\bibitem[Badamdorj et~al\mbox{.}(2021)]%
        {badamdorj2021joint}
\bibfield{author}{\bibinfo{person}{Taivanbat Badamdorj}, \bibinfo{person}{Mrigank Rochan}, \bibinfo{person}{Yang Wang}, {and} \bibinfo{person}{Li Cheng}.} \bibinfo{year}{2021}\natexlab{}.
\newblock \showarticletitle{Joint visual and audio learning for video highlight detection}. In \bibinfo{booktitle}{\emph{IEEE/CVF ICCV}}. \bibinfo{pages}{8127--8137}.
\newblock


\bibitem[Badamdorj et~al\mbox{.}(2022)]%
        {badamdorj2022contrastive}
\bibfield{author}{\bibinfo{person}{Taivanbat Badamdorj}, \bibinfo{person}{Mrigank Rochan}, \bibinfo{person}{Yang Wang}, {and} \bibinfo{person}{Li Cheng}.} \bibinfo{year}{2022}\natexlab{}.
\newblock \showarticletitle{Contrastive learning for unsupervised video highlight detection}. In \bibinfo{booktitle}{\emph{IEEE/CVF CVPR}}. \bibinfo{pages}{14042--14052}.
\newblock


\bibitem[Bateni{, \it et al.}(2020)]%
        {254354}
\bibfield{author}{\bibinfo{person}{S. Bateni{, \it et al.}}} \bibinfo{year}{2020}\natexlab{}.
\newblock \showarticletitle{{NeuOS}: A {Latency-Predictable} {Multi-Dimensional} Optimization Framework for {DNN-driven} Autonomous Systems}. In \bibinfo{booktitle}{\emph{USENIX ATC}}. \bibinfo{pages}{371--385}.
\newblock


\bibitem[big.LITTLE(2011)]%
        {big.little}
\bibfield{author}{\bibinfo{person}{big.LITTLE}.} \bibinfo{year}{2011}\natexlab{}.
\newblock \bibinfo{title}{Arm big.LITTLE heterogeneous processing architecture.}
\newblock
\newblock
\urldef\tempurl%
\url{https://www.arm.com/technologies/big-little/}
\showURL{%
\tempurl}


\bibitem[Cai et~al\mbox{.}(2021)]%
        {cai2021yolobile}
\bibfield{author}{\bibinfo{person}{Yuxuan Cai}, \bibinfo{person}{Hongjia Li}, \bibinfo{person}{Geng Yuan}, \bibinfo{person}{Wei Niu}, \bibinfo{person}{Yanyu Li}, \bibinfo{person}{Xulong Tang}, \bibinfo{person}{Bin Ren}, {and} \bibinfo{person}{Yanzhi Wang}.} \bibinfo{year}{2021}\natexlab{}.
\newblock \showarticletitle{Yolobile: Real-time object detection on mobile devices via compression-compilation co-design}. In \bibinfo{booktitle}{\emph{AAAI}}, Vol.~\bibinfo{volume}{35}. \bibinfo{pages}{955--963}.
\newblock


\bibitem[Girshick et~al\mbox{.}(2014)]%
        {girshick2014rich}
\bibfield{author}{\bibinfo{person}{Ross Girshick}, \bibinfo{person}{Jeff Donahue}, \bibinfo{person}{Trevor Darrell}, {and} \bibinfo{person}{Jitendra Malik}.} \bibinfo{year}{2014}\natexlab{}.
\newblock \showarticletitle{Rich feature hierarchies for accurate object detection and semantic segmentation}. In \bibinfo{booktitle}{\emph{IEEE/CVF CVPR}}. \bibinfo{pages}{580--587}.
\newblock


\bibitem[Glenn{, \it et al.}(2022)]%
        {yolov5}
\bibfield{author}{\bibinfo{person}{J. Glenn{, \it et al.}}} \bibinfo{year}{2022}\natexlab{}.
\newblock \bibinfo{booktitle}{\emph{ultralytics/yolov5: v7.0 - YOLOv5 SOTA Realtime Instance Segmentation}}.
\newblock
\urldef\tempurl%
\url{https://doi.org/10.5281/zenodo.7347926}
\showURL{%
\tempurl}


\bibitem[Gondimalla et~al\mbox{.}(2023)]%
        {10.1145/3613424.3614312}
\bibfield{author}{\bibinfo{person}{Ashish Gondimalla}, \bibinfo{person}{Mithuna Thottethodi}, {and} \bibinfo{person}{T.~N. Vijaykumar}.} \bibinfo{year}{2023}\natexlab{}.
\newblock \showarticletitle{Eureka: Efficient Tensor Cores for One-sided Unstructured Sparsity in DNN Inference}. In \bibinfo{booktitle}{\emph{Proceedings of the 56th Annual IEEE/ACM International Symposium on Microarchitecture}} (<conf-loc>, <city>Toronto</city>, <state>ON</state>, <country>Canada</country>, </conf-loc>) \emph{(\bibinfo{series}{MICRO '23})}. \bibinfo{publisher}{Association for Computing Machinery}, \bibinfo{address}{New York, NY, USA}, \bibinfo{pages}{324–337}.
\newblock
\showISBNx{9798400703294}
\urldef\tempurl%
\url{https://doi.org/10.1145/3613424.3614312}
\showDOI{\tempurl}


\bibitem[Gong et~al\mbox{.}(2022)]%
        {gong2022all}
\bibfield{author}{\bibinfo{person}{Yifan Gong}, \bibinfo{person}{Zheng Zhan}, \bibinfo{person}{Pu Zhao}, \bibinfo{person}{Yushu Wu}, \bibinfo{person}{Chao Wu}, \bibinfo{person}{Caiwen Ding}, \bibinfo{person}{Weiwen Jiang}, \bibinfo{person}{Minghai Qin}, {and} \bibinfo{person}{Yanzhi Wang}.} \bibinfo{year}{2022}\natexlab{}.
\newblock \showarticletitle{All-in-One: A Highly Representative DNN Pruning Framework for Edge Devices with Dynamic Power Management}. In \bibinfo{booktitle}{\emph{IEEE/ACM ICCAD}}. \bibinfo{pages}{1--9}.
\newblock


\bibitem[Gong{, \it et al.}(2023)]%
        {10247713}
\bibfield{author}{\bibinfo{person}{Y. Gong{, \it et al.}}} \bibinfo{year}{2023}\natexlab{}.
\newblock \showarticletitle{Condense: A Framework for Device and Frequency Adaptive Neural Network Models on the Edge}. In \bibinfo{booktitle}{\emph{ACM/IEEE DAC}}. \bibinfo{pages}{1--6}.
\newblock


\bibitem[Gowda et~al\mbox{.}(2021)]%
        {gowda2021smart}
\bibfield{author}{\bibinfo{person}{Shreyank~N Gowda}, \bibinfo{person}{Marcus Rohrbach}, {and} \bibinfo{person}{Laura Sevilla-Lara}.} \bibinfo{year}{2021}\natexlab{}.
\newblock \showarticletitle{Smart frame selection for action recognition}. In \bibinfo{booktitle}{\emph{AAAI}}, Vol.~\bibinfo{volume}{35}. \bibinfo{pages}{1451--1459}.
\newblock


\bibitem[Halpern{, \it et al.}(2016)]%
        {halpern2016mobile}
\bibfield{author}{\bibinfo{person}{M. Halpern{, \it et al.}}} \bibinfo{year}{2016}\natexlab{}.
\newblock \showarticletitle{Mobile CPU's rise to power: Quantifying the impact of generational mobile CPU design trends on performance, energy, and user satisfaction}. In \bibinfo{booktitle}{\emph{IEEE/ACM HPCA}}. \bibinfo{pages}{64--76}.
\newblock


\bibitem[Han{, \it et al.}(2021)]%
        {han2021herti}
\bibfield{author}{\bibinfo{person}{M. Han{, \it et al.}}} \bibinfo{year}{2021}\natexlab{}.
\newblock \showarticletitle{Herti: A reinforcement learning-augmented system for efficient real-time inference on heterogeneous embedded systems}. In \bibinfo{booktitle}{\emph{IEEE PACT}}. \bibinfo{pages}{90--102}.
\newblock


\bibitem[He et~al\mbox{.}(2017)]%
        {he2017mask}
\bibfield{author}{\bibinfo{person}{Kaiming He}, \bibinfo{person}{Georgia Gkioxari}, \bibinfo{person}{Piotr Doll{\'a}r}, {and} \bibinfo{person}{Ross Girshick}.} \bibinfo{year}{2017}\natexlab{}.
\newblock \showarticletitle{Mask r-cnn}. In \bibinfo{booktitle}{\emph{IEEE ICCV}}. \bibinfo{pages}{2961--2969}.
\newblock


\bibitem[Jiang and Mu(2022)]%
        {jiang2022joint}
\bibfield{author}{\bibinfo{person}{Hao Jiang} {and} \bibinfo{person}{Yadong Mu}.} \bibinfo{year}{2022}\natexlab{}.
\newblock \showarticletitle{Joint video summarization and moment localization by cross-task sample transfer}. In \bibinfo{booktitle}{\emph{IEEE/CVF CVPR}}. \bibinfo{pages}{16388--16398}.
\newblock


\bibitem[Kim{, \it et al.}(2021)]%
        {kim2021ztt}
\bibfield{author}{\bibinfo{person}{S. Kim{, \it et al.}}} \bibinfo{year}{2021}\natexlab{}.
\newblock \showarticletitle{zTT: Learning-based DVFs with zero thermal throttling for mobile devices}. In \bibinfo{booktitle}{\emph{MobiSys}}. \bibinfo{pages}{41--53}.
\newblock


\bibitem[Li et~al\mbox{.}(2023)]%
        {li2023automated}
\bibfield{author}{\bibinfo{person}{M. Li} {et~al\mbox{.}}} \bibinfo{year}{2023}\natexlab{}.
\newblock \showarticletitle{Automated Optical Accelerator Search Toward Superior Acceleration Efficiency, Inference Robustness and Development Speed}.
\newblock \bibinfo{journal}{\emph{IEEE TCAD}} (\bibinfo{year}{2023}).
\newblock


\bibitem[Li{, \it et al.}(2020)]%
        {li2020reducto}
\bibfield{author}{\bibinfo{person}{Y. Li{, \it et al.}}} \bibinfo{year}{2020}\natexlab{}.
\newblock \showarticletitle{Reducto: On-camera filtering for resource-efficient real-time video analytics}. In \bibinfo{booktitle}{\emph{SIGCOMM}}. \bibinfo{pages}{359--376}.
\newblock


\bibitem[Liu{, \it et al.}(2020)]%
        {liu2020cocopie}
\bibfield{author}{\bibinfo{person}{S. Liu{, \it et al.}}} \bibinfo{year}{2020}\natexlab{}.
\newblock \showarticletitle{CoCoPIE: Making Mobile AI Sweet As PIE--Compression-Compilation Co-Design Goes a Long Way}.
\newblock \bibinfo{journal}{\emph{arXiv}} (\bibinfo{year}{2020}).
\newblock


\bibitem[Liu{, \it et al.}(2016)]%
        {liu2016ssd}
\bibfield{author}{\bibinfo{person}{W. Liu{, \it et al.}}} \bibinfo{year}{2016}\natexlab{}.
\newblock \showarticletitle{Ssd: Single shot multibox detector}. In \bibinfo{booktitle}{\emph{ECCV}}. \bibinfo{pages}{21--37}.
\newblock


\bibitem[Ma{, \it et al.}(2022)]%
        {ma2022blcr}
\bibfield{author}{\bibinfo{person}{X. Ma{, \it et al.}}} \bibinfo{year}{2022}\natexlab{}.
\newblock \showarticletitle{Blcr: Towards real-time dnn execution with block-based reweighted pruning}. In \bibinfo{booktitle}{\emph{IEEE ISQED}}. \bibinfo{pages}{1--8}.
\newblock


\bibitem[Narasimhan et~al\mbox{.}(2021)]%
        {narasimhan2021clip}
\bibfield{author}{\bibinfo{person}{Medhini Narasimhan}, \bibinfo{person}{Anna Rohrbach}, {and} \bibinfo{person}{Trevor Darrell}.} \bibinfo{year}{2021}\natexlab{}.
\newblock \showarticletitle{Clip-it! language-guided video summarization}.
\newblock \bibinfo{journal}{\emph{NIPS}}  \bibinfo{volume}{34} (\bibinfo{year}{2021}), \bibinfo{pages}{13988--14000}.
\newblock


\bibitem[Niu{, \it et al.}(2020)]%
        {niu2020patdnn}
\bibfield{author}{\bibinfo{person}{W. Niu{, \it et al.}}} \bibinfo{year}{2020}\natexlab{}.
\newblock \showarticletitle{Patdnn: Achieving real-time dnn execution on mobile devices with pattern-based weight pruning}. In \bibinfo{booktitle}{\emph{ASPLOS}}. \bibinfo{pages}{907--922}.
\newblock


\bibitem[{R Core Team}(2021)]%
        {Moonson}
\bibfield{author}{\bibinfo{person}{{R Core Team}}.} \bibinfo{year}{2021}\natexlab{}.
\newblock \bibinfo{booktitle}{\emph{Moonson High Voltage Power Monitor}}.
\newblock Moonson Solutions Inc.
\newblock
\urldef\tempurl%
\url{https://www.msoon.com/online-store/}
\showURL{%
\tempurl}


\bibitem[Ren{, \it et al.}(2015)]%
        {ren2015faster}
\bibfield{author}{\bibinfo{person}{S. Ren{, \it et al.}}} \bibinfo{year}{2015}\natexlab{}.
\newblock \showarticletitle{Faster r-cnn: Towards real-time object detection with region proposal networks}.
\newblock \bibinfo{journal}{\emph{NIPS}}  \bibinfo{volume}{28} (\bibinfo{year}{2015}).
\newblock


\bibitem[Ross(2015)]%
        {girshick2015fast}
\bibfield{author}{\bibinfo{person}{G. Ross}.} \bibinfo{year}{2015}\natexlab{}.
\newblock \showarticletitle{Fast r-cnn}. In \bibinfo{booktitle}{\emph{ICCV}}. \bibinfo{pages}{1440--1448}.
\newblock


\bibitem[Roychowdhury(2022)]%
        {roychowdhury2022semi}
\bibfield{author}{\bibinfo{person}{S. Roychowdhury}.} \bibinfo{year}{2022}\natexlab{}.
\newblock \showarticletitle{Semi-supervised and Deep learning Frameworks for Video Classification and Key-frame Identification}. In \bibinfo{booktitle}{\emph{IEEE IJCNN}}. \bibinfo{pages}{1--8}.
\newblock


\bibitem[Song{, \it et al.}(2021)]%
        {song2021dancing}
\bibfield{author}{\bibinfo{person}{Y. Song{, \it et al.}}} \bibinfo{year}{2021}\natexlab{}.
\newblock \showarticletitle{Dancing along battery: Enabling transformer with run-time reconfigurability on mobile devices}. In \bibinfo{booktitle}{\emph{ACM/IEEE DAC}}. \bibinfo{pages}{1003--1008}.
\newblock


\bibitem[Tang et~al\mbox{.}(2023)]%
        {tang2023deep}
\bibfield{author}{\bibinfo{person}{Hao Tang}, \bibinfo{person}{Lei Ding}, \bibinfo{person}{Songsong Wu}, \bibinfo{person}{Bin Ren}, \bibinfo{person}{Nicu Sebe}, {and} \bibinfo{person}{Paolo Rota}.} \bibinfo{year}{2023}\natexlab{}.
\newblock \showarticletitle{Deep unsupervised key frame extraction for efficient video classification}.
\newblock \bibinfo{journal}{\emph{ACM TOMM}} \bibinfo{volume}{19}, \bibinfo{number}{3} (\bibinfo{year}{2023}), \bibinfo{pages}{1--17}.
\newblock


\bibitem[Tang{, \it et al.}(2019)]%
        {tang2019impact}
\bibfield{author}{\bibinfo{person}{Z. Tang{, \it et al.}}} \bibinfo{year}{2019}\natexlab{}.
\newblock \showarticletitle{The impact of GPU DVFS on the energy and performance of deep learning: An empirical study}. In \bibinfo{booktitle}{\emph{ACM e-Energy}}. \bibinfo{pages}{315--325}.
\newblock


\bibitem[Umme{, \it et al.}(2019)]%
        {sara2019image}
\bibfield{author}{\bibinfo{person}{S. Umme{, \it et al.}}} \bibinfo{year}{2019}\natexlab{}.
\newblock \showarticletitle{Image quality assessment through FSIM, SSIM, MSE and PSNR—a comparative study}.
\newblock \bibinfo{journal}{\emph{Journal of Computer and Communications}} \bibinfo{volume}{7}, \bibinfo{number}{3} (\bibinfo{year}{2019}), \bibinfo{pages}{8--18}.
\newblock


\bibitem[Wang et~al\mbox{.}(2023)]%
        {wang2023yolov7}
\bibfield{author}{\bibinfo{person}{Chien-Yao Wang}, \bibinfo{person}{Alexey Bochkovskiy}, {and} \bibinfo{person}{Hong-Yuan~Mark Liao}.} \bibinfo{year}{2023}\natexlab{}.
\newblock \showarticletitle{YOLOv7: Trainable bag-of-freebies sets new state-of-the-art for real-time object detectors}. In \bibinfo{booktitle}{\emph{Proceedings of the IEEE/CVF conference on computer vision and pattern recognition}}. \bibinfo{pages}{7464--7475}.
\newblock


\bibitem[Wang{, \it et al.}(2021)]%
        {wang2021asymo}
\bibfield{author}{\bibinfo{person}{M. Wang{, \it et al.}}} \bibinfo{year}{2021}\natexlab{}.
\newblock \showarticletitle{Asymo: scalable and efficient deep-learning inference on asymmetric mobile cpus}. In \bibinfo{booktitle}{\emph{MobiCom}}. \bibinfo{pages}{215--228}.
\newblock


\bibitem[Wang~Zhou(2004)]%
        {wang2004image}
\bibfield{author}{\bibinfo{person}{et~al. Wang~Zhou}.} \bibinfo{year}{2004}\natexlab{}.
\newblock \showarticletitle{Image quality assessment: from error visibility to structural similarity}.
\newblock \bibinfo{journal}{\emph{IEEE T-IP}} \bibinfo{volume}{13}, \bibinfo{number}{4} (\bibinfo{year}{2004}), \bibinfo{pages}{600--612}.
\newblock


\bibitem[Wen et~al\mbox{.}(2016)]%
        {wen2016learning}
\bibfield{author}{\bibinfo{person}{Wei Wen}, \bibinfo{person}{Chunpeng Wu}, \bibinfo{person}{Yandan Wang}, \bibinfo{person}{Yiran Chen}, {and} \bibinfo{person}{Hai Li}.} \bibinfo{year}{2016}\natexlab{}.
\newblock \showarticletitle{Learning structured sparsity in deep neural networks}.
\newblock \bibinfo{journal}{\emph{NIPS}}  \bibinfo{volume}{29} (\bibinfo{year}{2016}).
\newblock


\bibitem[Wu et~al\mbox{.}(2022)]%
        {wu2022compiler}
\bibfield{author}{\bibinfo{person}{Yushu Wu}, \bibinfo{person}{Yifan Gong}, \bibinfo{person}{Pu Zhao}, \bibinfo{person}{Yanyu Li}, \bibinfo{person}{Zheng Zhan}, \bibinfo{person}{Wei Niu}, \bibinfo{person}{Hao Tang}, \bibinfo{person}{Minghai Qin}, \bibinfo{person}{Bin Ren}, {and} \bibinfo{person}{Yanzhi Wang}.} \bibinfo{year}{2022}\natexlab{}.
\newblock \showarticletitle{Compiler-aware neural architecture search for on-mobile real-time super-resolution}. In \bibinfo{booktitle}{\emph{Springer ECCV}}. \bibinfo{pages}{92--111}.
\newblock


\bibitem[Wu{, \it et al.}(2020)]%
        {wu2020pruning}
\bibfield{author}{\bibinfo{person}{C. Wu{, \it et al.}}} \bibinfo{year}{2020}\natexlab{}.
\newblock \showarticletitle{Pruning deep reinforcement learning for dual user experience and storage lifetime improvement on mobile devices}.
\newblock \bibinfo{journal}{\emph{IEEE TCAD}} \bibinfo{volume}{39}, \bibinfo{number}{11} (\bibinfo{year}{2020}), \bibinfo{pages}{3993--4005}.
\newblock


\bibitem[Wu{, \it et al.}(2023)]%
        {wu2023iccad}
\bibfield{author}{\bibinfo{person}{Y. Wu{, \it et al.}}} \bibinfo{year}{2023}\natexlab{}.
\newblock \showarticletitle{MOC: Multi-Objective Mobile CPU-GPU Co-optimization for Power-efficient DNN Inference}. In \bibinfo{booktitle}{\emph{ICCAD}}.
\newblock


\bibitem[Yu et~al\mbox{.}(2018)]%
        {yu2018slimmable}
\bibfield{author}{\bibinfo{person}{Jiahui Yu}, \bibinfo{person}{Linjie Yang}, \bibinfo{person}{Ning Xu}, \bibinfo{person}{Jianchao Yang}, {and} \bibinfo{person}{Thomas Huang}.} \bibinfo{year}{2018}\natexlab{}.
\newblock \showarticletitle{Slimmable neural networks}.
\newblock \bibinfo{journal}{\emph{arXiv preprint arXiv:1812.08928}} (\bibinfo{year}{2018}).
\newblock


\bibitem[Yu{, \it et al.}(2020)]%
        {yu2020bdd100k}
\bibfield{author}{\bibinfo{person}{F. Yu{, \it et al.}}} \bibinfo{year}{2020}\natexlab{}.
\newblock \showarticletitle{Bdd100k: A diverse driving dataset for heterogeneous multitask learning}. In \bibinfo{booktitle}{\emph{IEEE/CVF CVPR}}. \bibinfo{pages}{2636--2645}.
\newblock


\bibitem[Yuan{, \it et al.}(2021)]%
        {yuan2021mest}
\bibfield{author}{\bibinfo{person}{G. Yuan{, \it et al.}}} \bibinfo{year}{2021}\natexlab{}.
\newblock \showarticletitle{Mest: Accurate and fast memory-economic sparse training framework on the edge}.
\newblock \bibinfo{journal}{\emph{NIPS}}  \bibinfo{volume}{34} (\bibinfo{year}{2021}), \bibinfo{pages}{20838--20850}.
\newblock


\bibitem[Zhang{, \it et al.}(2019)]%
        {zhang2019determinants}
\bibfield{author}{\bibinfo{person}{B. Zhang{, \it et al.}}} \bibinfo{year}{2019}\natexlab{}.
\newblock \showarticletitle{Determinants of take-over time from automated driving: A meta-analysis of 129 studies}.
\newblock \bibinfo{journal}{\emph{Transportation research part F: traffic psychology and behavior}}  \bibinfo{volume}{64} (\bibinfo{year}{2019}), \bibinfo{pages}{285--307}.
\newblock


\bibitem[Zhong et~al\mbox{.}(2021)]%
        {zhong2021revisit}
\bibfield{author}{\bibinfo{person}{Shaochen Zhong}, \bibinfo{person}{Guanqun Zhang}, \bibinfo{person}{Ningjia Huang}, {and} \bibinfo{person}{Shuai Xu}.} \bibinfo{year}{2021}\natexlab{}.
\newblock \showarticletitle{Revisit kernel pruning with lottery regulated grouped convolutions}. In \bibinfo{booktitle}{\emph{ICLR}}.
\newblock


\end{thebibliography}


\end{document}